\documentclass[sigconf]{acmart}
\AtBeginDocument{%
  \providecommand\BibTeX{{%
    \normalfont B\kern-0.5em{\scshape i\kern-0.25em b}\kern-0.8em\TeX}}}

\usepackage{multicol,multirow}
\usepackage{subfigure}
\usepackage{graphicx}

\begin{document}


\title{On the Use of Time Series Kernel and Dimensionality Reduction to Identify the Acquisition of Antimicrobial Multidrug Resistance in the Intensive Care Unit}

\author{Óscar Escudero-Arnanz}
\email{oscar.escudero@urjc.es}
\affiliation{%
  \institution{Rey Juan Carlos University}
  \streetaddress{Camino del Molino, 5}
  \city{Fuenlabrada}
  \state{Madrid}
  \country{Spain}
  \postcode{28942}
}

\author{Joaquín Rodríguez-Álvarez}
\email{joaquin.alvarez@salud.madrid.org}
\affiliation{%
  \institution{University Hospital of Fuenlabrada}
  \streetaddress{Camino del Molino, 2}
   \city{Fuenlabrada}
  \state{Madrid}
  \country{Spain}
  \postcode{28942}
}

\author{Karl {\O}yvind Mikalsen}
\email{karl.o.mikalsen@uit.no}
\affiliation{%
 \institution{University Hospital of North-Norway, UiT  The Arctic University of Norway}
 \streetaddress{Sykehusvegen, 38}
 \city{Troms{\o}}
 \country{Norway}
}

\author{Robert Jenssen}
\email{robert.jenssen@uit.no}
\affiliation{%
  \institution{UiT The Arctic University of Norway}
  \streetaddress{Hansine Hansens veg 18}
  \city{Troms{\o}}
  \country{Norway}}

\author{Cristina Soguero-Ruiz}
\email{cristina.soguero@urjc.es}
\affiliation{%
  \institution{Rey Juan Carlos University}
  \streetaddress{Camino del Molino, 5}
  \city{Fuenlabrada}
  \state{Madrid}
  \country{Spain}
  \postcode{28942}
}


\renewcommand{\shortauthors}{Escudero-Arnanz, et al.}

\begin{abstract}

The acquisition of Antimicrobial Multidrug Resistance (AMR) in patients admitted to the Intensive Care Units (ICU) is a major global concern. This study analyses data in the form of multivariate time series (MTS) from 3476 patients recorded at the ICU of University Hospital of Fuenlabrada (Madrid) from 2004 to 2020. 18\% of the patients acquired AMR during their stay in the ICU. The goal of this paper is an early prediction of the development of AMR. Towards that end, we leverage the time-series cluster kernel (TCK) to learn similarities between MTS. To evaluate the effectiveness of TCK as a kernel, we applied several dimensionality reduction techniques for visualization and classification tasks. The experimental results show that TCK allows identifying a group of patients that acquire the AMR during the first 48 hours of their ICU stay, and it also provides good classification capabilities.

\end{abstract}





\maketitle

\section{Introduction}

Longitudinal Electronic Health Records (EHR), which thoroughly collect patient health information over time, have proven to be one of the most relevant data sources for tasks such as early prediction of anastomosis leakage~\cite{soguero2016predicting}, characterization of patient health-status~\cite{chushig2020data}, and prediction of type 2 diabetes~\cite{garcia2020use}. However, many challenges have been raised when analyzing temporal EHR-based data. Such multivariate time series (MTS) can be characterized by missing values, different length and possibly dependent variables~\cite{mikalsen2018time}. To deal with these issues, several methods have been proposed to exploit temporal clinical data~\cite{mikalsen2018time}. Among them, we explore the potential of the time-series cluster kernel (TCK), which computes the pairwise similarities between time series with missing data. The created kernel matrix can be used for many different purposes, such as dimensionality reduction (DR) or classification.


Learning compressed representations of MTS make data analysis easier in the presence of redundant data, as well as for a high number of variables and time steps. Traditional DR algorithms are designed for vectorial data. However, in this paper, we leverage the potential of TCK to map high-dimensional into much lower-dimensional space. Towards that end, representing learning, i.e., transforming the input space to a new feature representation space by linear and non-linear approaches, are considered. The learning compressed representations of MTS 
can be used to identify visually patients with specific clinical characteristics. On the other hand, this new space can be considered as the input space for linear and non-linear classifiers.

The described methodology is applied in this work to identify the acquisition of antimicrobial multidrug resistance (AMR) in the Intensive Care Unit (ICU). This is a growing problem that jeopardizes seven decades of medical progress since antibiotics were first used in clinical practice~\cite{world2014antimicrobial}. The misuse and overuse of antibiotics have resulted in bacteria being resistant to one or more antibiotics, no longer responding to drugs that they were initially sensitive to. The lack of antimicrobial effectiveness could increase the risk when treating infections, becoming impossible or extremely difficult to find a suitable treatment to cure them~\cite{world2014antimicrobial}. This situation is even more critical in the ICU due to the delicate health condition of the patients in this unit.

As a consequence, AMR is causing a significant social and economic burden worldwide~\cite{GlobalActionPlan}. Antibiotic resistance is estimated to be responsible for nearly 300 million premature deaths and considerable economic losses by 2050, according to a recent study~\cite{munita2016mechanisms}. The overall economic cost of AMR was predicted to be approximately 1.5 billion euros, with hospital expenditures accounting for 900 million~\cite{prestinaci2015antimicrobial}. This paper, therefore, proposes an approach to earlier identify the development and spread of AMR in the ICU. Towards that end, MTS associated with the use of antibiotics in this unit are analyzed. 

The structure of this paper is as follows. Section 2 provides an overview of the data and the methods used in the paper. Section 3 presents the experimental results, whereas discussion and conclusions are included in Section 4.


\section{Data and methods}

\subsection{Data}

The dataset used in the current study consisted of MTS extracted from the EHR of the ICU at the University Hospital of Fuenlabrada from 2004 until 2020. From 3476 patients admitted to the ICU during that period, 628 patients developed AMR. Each patient is characterized by MTS related to the family of antibiotics taken by a specific patient during his/her ICU stay, as well as the antibiotics taken by patients who shared the clinical unit during the stay of the patient to be studied. Moreover, we count the number of patients who shared the clinical unit and the number of AMR patients at a given time (24 hours slot). We also analyze if the patient has been assisted with mechanical ventilation. The family of antibiotics considered in this work are: Aminoglycosides (AMG), Antifungals (ATF), Carbapenemes (CAR), 1st generation Cephalosporins (CF1), 2nd generation Cephalosporins (CF2), 3rd generation Cephalosporins (CF3), 4th generation Cephalosporins (CF4), unclassified antibiotics (Others), Glycyclines (GCC), Glycopeptides (GLI), Lincosamides (LIN), Lipopeptides (LIP), Macrolides (MAC), Monobactamas (MON), Nitroimidazolics (NTI), Miscellaneous (OTR), Oxazolidinones (OXA), Broad-Spectrum Penicillins (PAP), Penicillins (PEN), Polypeptides (POL), Quinolones (QUI), Sulfamides (SUL) and Tetracyclines (TTC).

On average, the first multidrug resistance is detected within seven days after patient admission to ICU, similar to the average length of stay of non-AMR patients. Based on these results, we determine to be seven days the length of the longest MTS. Therefore, we fill with zero values the time observation of patients whose stays in the ICU were less than seven days. If the length of stay is longer than seven days, we consider the information corresponding to the last seven days closest to the detection of the first AMR. For non-AMR patients, and based on clinical knowledge, the patient’s admission to the ICU is the reference (see Figure 1 for details).

Hence, the dataset is represented as $D=\{\mathbf{X}_{i}, y_{i}\}_{i=1}^n $, where the $i$-th patient is represented by the temporal matrix $\mathbf{X}_i$ and the output $y_{i}$, which identifies if a patient acquired (``1'') or not (``0'') an AMR during his/her stay in the ICU. The matrix $\mathbf{X}_i$ modelled for the $i$-th patient $D$ time series, each of them defined by a number of observations $T$, as follows: $\mathbf{X}_i~=~[\mathbf{x}_{i}^{(1)},\ldots,\mathbf{x}_{i}^{(T)} ]$, $\in$ $\mathbb{R}^{D \times T}$, with the column vector $\mathbf{x}_i^{(t)}$ having length $D$ for all $i$ and $t$. 

\begin{figure}[h]
    \centering
    \includegraphics[width=0.5\textwidth]{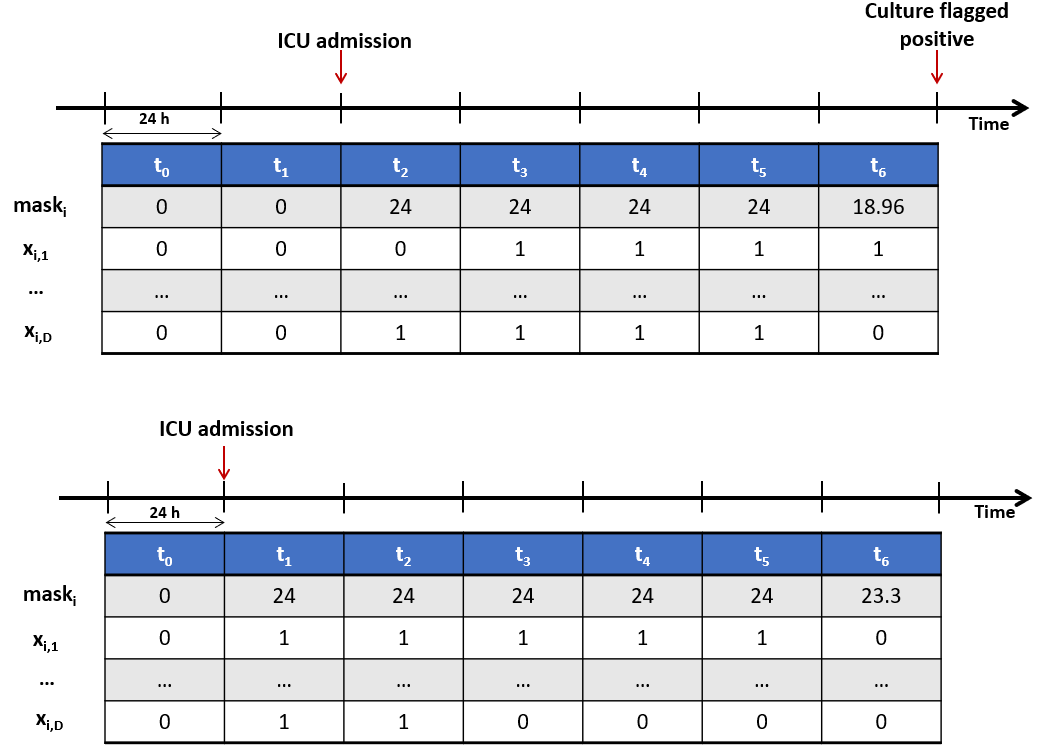}
    \caption{Schema of the 7-days time window considered for AMR patients (upper panel) and non-AMR patients (bottom panel).}
    \label{TemporalWindow}
\end{figure}

\subsection{Methods}



MTS have been analyzed in a variety of applications such as financial or health~\cite{chatfield2003analysis}.  From a theoretical point of view, several studies have considered a classical approach aiming to deal with MTS by extracting handcrafted features from raw data~\cite{soguero2015data,soguero2016predicting}. Others have focused on computing the pairwise learning similarities between the time series, such as dynamic time warping (DTW)~\cite{wang2013experimental,mikalsen2021time}.  However, many are not suitable for kernel methods due to not satisfying the condition of being positive semi-definite. 

A method known as time series cluster kernel (TCK) is employed in this study. This method is based on ensemble learning approaches and probabilistic models known as Gaussian Models (GMMs). GMMs are fitting to a randomly chosen subset of MTS, features and time segments by considering different numbers of mixture components and random initial conditions. To estimate model parameters (time-dependent means, covariance matrix, and the variance of the attribute) when dealing with missing data, the likelihoods are multiplied with informative priors for
the parameters, and maximum a posteriori expectation-maximization is considered~\cite{marlin2012unsupervised}. After convergence, the posterior probability of each GMM is obtained.  The inner products between pairs of posterior probabilities provided by each partition are summing up to build the kernel matrix, following the ensemble strategy.  Therefore, given a GMM ensemble, we compute the TCK by exploiting the fact that the sum of kernels is itself a kernel. Since TCK procedure generates partitions at different resolutions that capture both local and global structures in the data, it can capture local and global relationships in the underlying data, it is robust to outliers and parameter-free. More details on the TCK are provided in~\cite{mikalsen2018time}. We evaluate the potential of the learned representations (kernel) for dimensionality reduction, visualization and classification tasks.

Regarding dimensionality reduction, we focus on linear and non-linear dimensionality reduction methods to represent the embedding of the EHR MTS in the TCK space. Principal Component Analysis (PCA) is considered to explore the linear transformations~\cite{anowar2021conceptual}, whereas kernel PCA (KPCA) and autoencoders (AE) are considered as non-linear dimensionality reduction approaches. Note that AE are used to learn data representations in deep architectures, see~\cite{vincent2008extracting} for more details. To visualize data in two dimensions, we apply t-Distributed Stochastic Neighbor Embedding (t-SNE)~\cite{van2008visualizing}.


Regarding classification, the learning representation is used as the input to different classifiers. In this work, we apply linear (Logistic Regression, LR) and non-linear classifiers (k-nearest neighbour, k-NN; decision trees; random forest; support vector machines, SVM; nu-SVM; and multilayer perceptron, MLP). Due to space limitations, we do not describe the classifiers here, but for the interested reader, we refer to~\cite{bishop2006pattern}.

All experiments were performed using Python language, and to model the AE, we used Keras.

\section{Results}

\begin{figure}[H]
    \centering
	\includegraphics[width=0.47\textwidth]{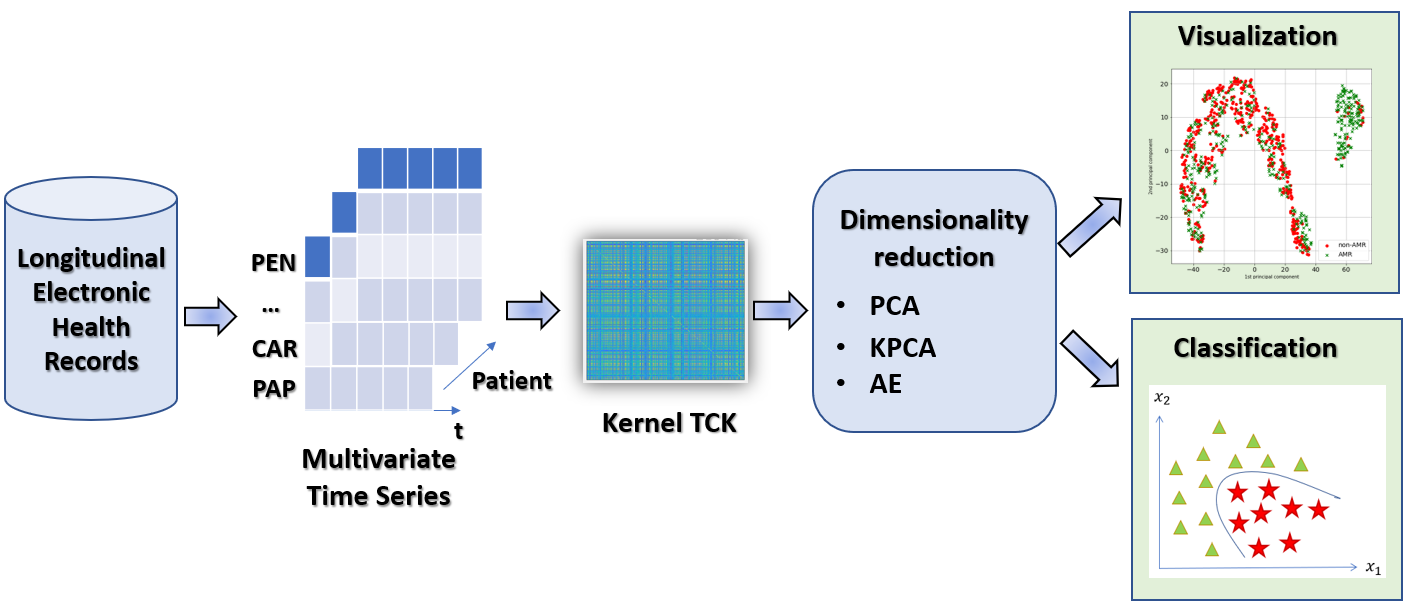}
    \caption{Schematic description of the methodology followed in this paper.}
    \label{Resum}
\end{figure}

This section aims to evaluate the effectiveness of the TCK  by applying different dimensionality reduction techniques: PCA, KPCA and AE. After using these methods, the resulting learning representations are used for 2D visualization using t-SNE and for classification purposes. A summary of the process followed in this work is shown in Figure \ref{Resum}.
The original dataset is separated into two subsets, training and test, which account for 70\% and 30\% of the patients, respectively~\cite{caruana2015intelligible}. The train set is balanced concerning the minority class (AMR-patients), using the remaining data in the test set (non-AMR patients). We apply the TCK to this dataset (freely available Matlab code in \cite{codeMikalsen}), considering the maximum number of mixtures component for each Gaussian Mixture Models to be 40, and the number of randomizations for each number of components equals 30.



\subsubsection*{\textbf{Dimensionality reduction and visualization.}} To visually evaluate the potential of TCK as a kernel when dealing with MTS, we benchmark PCA with TCK, KPCA with TCK and AE with TCK. Note that for PCA, we decide to capture 99\% of the information of the original space, ending up in 16 principal components. For KPCA, we consider a polynomial kernel, 50 principal components and a gamma value of 0.002083. These hyperparameters are tuned based on the minimum mean square error between the original and the compressed space obtained in the validation set.  The same criteria are applied for AE, for which a leakyRelu activation function is used, except for the last layer, where a sigmoid is considered. The minimum mean squared error was used as the loss function. The AE is trained for 1000 epochs with an Adam optimizer and exponential learning rate decay. Several simple and deep AE are evaluating, showing that considering 712 hidden neurons and 250 neurons in the compressed space is the best architecture to identify AMR patients. Keras in Tensorflow has been used for this implementation.

\begin{figure*}[h!]
\begin{subfigure}[]
	\centering
		\includegraphics[width=0.32\textwidth]{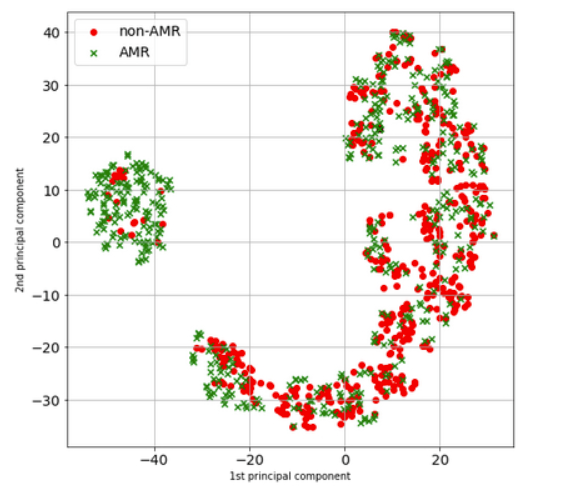}
	\end{subfigure}
	\begin{subfigure}[]
		\centering
		\includegraphics[width=0.32\textwidth]{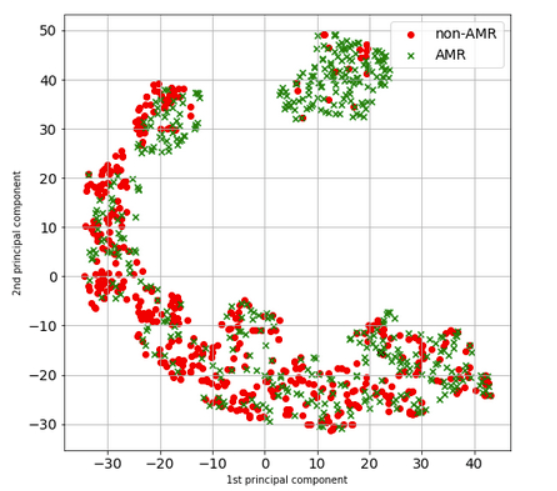}
	\end{subfigure}
	\begin{subfigure}[]
		\centering
		\includegraphics[width=0.32\textwidth]{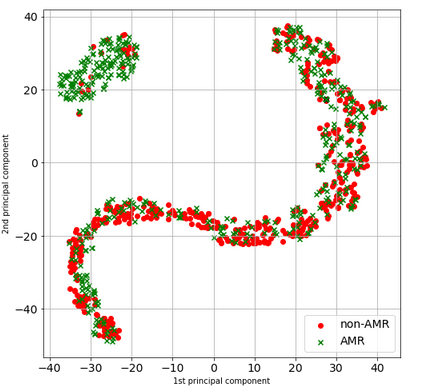}
	\end{subfigure}
	\centering
	 \caption{Visualization  of AMR and non-AMR patients using t-SNE representations after reduction of the TCK space with PCA (first column), KPCA (second column) and AE (third column).
}
	 \label{fig:results2D}
\end{figure*}

\begin{figure}[t!]
    \centering
    \begin{subfigure}[]
	\centering
		\includegraphics[width=0.45\textwidth]{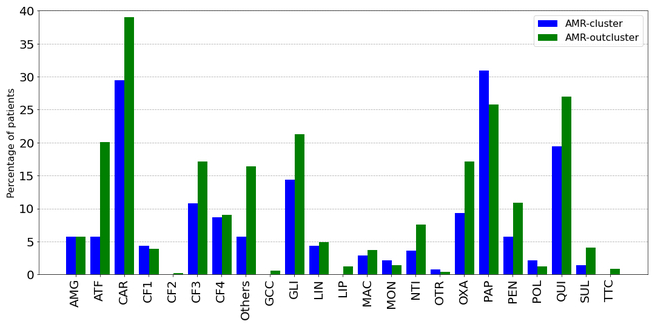}
	\end{subfigure}
	\begin{subfigure}[]
		\centering
		\includegraphics[width=0.45\textwidth]{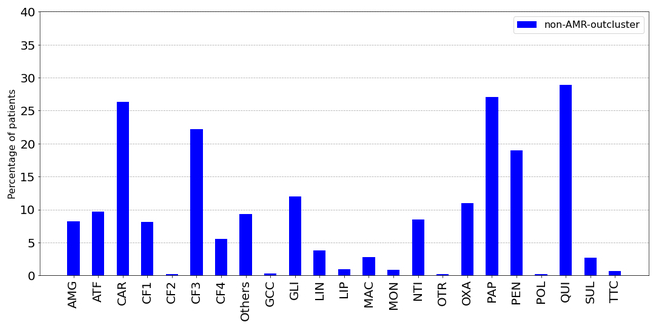}
	\end{subfigure}
    \caption{Percentage of AMR patients within and out of the cluster (a); and percentage of non-AMR patients (b).}
    \label{Antib}
\end{figure}






The new representations spaces are considered as input to t-SNE, aiming to visualize patients in two dimensions. These visualizations are shown in Figure~\ref{fig:results2D} for (a) for PCA, in  
 Figure \ref{fig:results2D} (b) for KPCA, and in Figure \ref{fig:results2D} (c) for AE. The learning representations provides knowledge for AMR patient identification. 
A distinguishable cluster (colored mainly in green) is observed in Figures \ref{fig:results2D} (a), (b) and (c), composed by 157, 157 and 161 patients, respectively. The patients grouped in the cluster in Figures \ref{fig:results2D} (a) and (b), are part of the patients observed in the cluster shown in Figure~\ref{fig:results2D} (c). 
It is important to highlight that, in this cluster, the majority (139) are patients with AMR detected in the first 48 hours of ICU stay, of whom 61.87\% required mechanical ventilation, compared to 76.68\% of AMR patients not within the cluster. In this line, AMR patients outside the cluster require more antibiotic treatments (see Figure~\ref{Antib} (a) for details). This may support that their health status is more critical.  Furthermore, it can be observed that, in general, non-AMR patients take fewer antibiotics than AMR patients, except for families of antibiotics such as PEN and CF3 (see Figure~\ref{Antib} (a) and (b) for details).

\subsubsection*{\textbf{Classification results.}}  The learned representations by PCA, KPCA and AE are used as the input of different linear and non-linear classifiers, specifically, LR, k-NN, decision tree, random forests, SVM, nu-SVM and MLP. The metrics used to measure the performance of the classifiers are accuracy, specificity, sensitivity, and area under the curve (AUC). To tune the hyperparameters, a 5-fold cross-validation strategy was considered in the training set. Results in the test set are shown in Table~\ref{tab:result}. Note that, in general, AE is the most adequate DR approach. It can also be observed that linear classifiers perform well in terms of sensitivity and AUC, whereas non-linear classifiers, such as nu-SVM, provide better accuracy and specificity results.

\begin{table}[t!]
  \caption{Classification results provided for different DR and classifier in terms of accuracy, specificity, sensitivity and AUC.}
  \label{tab:result}
  \resizebox{8,7cm}{!}{
  \begin{tabular}{c|c|c|c|c|c}
    \toprule
    \textbf{DR Method} & \textbf{Classifier}  & \textbf{Accuracy} & \textbf{Specificity} & \textbf{Sensitivity} & \textbf{AUC}\\
    \midrule
    \multirow{7}{*}{\textbf{PCA}} 
    & LR  & 53.49 & 51.73 & 76.8 & 64.26 \\
    & k-NN  & 64.91 & 64.31 & 72.93 & 68.62\\ 
    & Tree  & 57.47 & 58.64 & 41.99 & 50.32 \\
    & Random forest  & 65.92 & 67.47 & 45.3 & 56.39 \\
    & nu-SVM & 57.71 & 56.77 & 70.17 & 63.47 \\
    & SVM &  58.48 & 57.43 & 72.38 & 64.91 \\
    & MLP  & 51.74 & 49.65 & 79.55 & 64.60 \\
    \midrule
    \multirow{7}{*}{\textbf{KPCA}} 
    & LR & 51.7 & 50.06 & 73.48 & 61.77 \\
    & k-NN & 62.97 & 62.52 & 69.06 & 65.79 \\
    & Tree  & 58.46 & 60.43 & 32.23 & 46.33 \\
    & Random forest  & 70.95 & 71.68 & 61.33 & 66.5 \\
    & nu-SVM  & 57.32 & 58.31 & 44.2 & 51.25 \\
    & SVM  & 54.8 & 53.44 & 72.93 & 63.18 \\
    & MLP & 55.61 & 54.47 & 70.71 & 62.59 \\
    \midrule
    \multirow{7}{*}{\textbf{AE}} 
    & LR  & 82.26 & 82.8 & \textbf{75.14} & \textbf{78.97} \\
    & k-NN  & 79.86 & 81.8 & 54.14 & 67.97 \\
    & Tree  & 75.95 & 77.01 & 61.88 & 69.44 \\
    & Random forest  & 77.85 & 79.13 & 60.77 & 69.95 \\
    & nu-SVM  & \textbf{82.92} & \textbf{83.59} & 74.03 & 78.81 \\
    & SVM  & 75.99 & 76.51 & 69.06 & 72.79 \\
    & MLP  & 80.94 & 81.92 & 67.95 & 74.93 \\
  \bottomrule
\end{tabular}
}
\end{table}


\section{Discussion and conclusion}

This work presents a promising approach for early identification of AMR  patients in the ICU based on MTS recorded in EHR. The following are some of our contributions:
\begin{itemize}
    \item A time series cluster method is created to find similarity measures for MTS with missing data.
    \item Compressed representations that preserve pairwise relationships allows clinicians to visually identify the acquisition of AMR in the ICU.
    \item Classification results considering the learning representation as input space suggest that the proposed methodology can be used for earlier detection of AMR.
\end{itemize}

Learning compressed representations of the TCK space based on linear and non-linear approaches provides promising visualization for identifying a specific group of AMR patients who acquired the AMR during the first 48 hours of their stay in the ICU. This allows anticipating the culture results and taking isolation measures to avoid further spreading to other patients in the unit. The experimental results also provide good classification capabilities, bringing some light to the antibiotic treatment used to treat AMR patients.

The potential of deep autoencoders in this study opens the way for exploring more complex AE such as denoising or variational autoencoders~\cite{doersch2016tutorial}. 
Future work also includes the possibility of considering this problem as a multiclass classification problem rather than a binary one, aiming to distinguish between AMR detected in the first 48 hours, AMR detected later and non-AMR patients.

\section{ACKNOWLEDGMENTS}
This work has been partly supported by the Spanish Research projects PID2019-107768RA-I00 (AAVis-BMR), PID2019-106623RB-C41 (Beyond), DTS17/00158, and Project Ref. F661 (Mapping-UCI)- by the Community of Madrid and the Rey Juan Carlos University.
\bibliographystyle{ACM-Reference-Format}
\bibliography{sample-base}

\end{document}